\documentclass[10pt,twocolumn,letterpaper]{article}

 \usepackage[pagenumbers]{cvpr} 

\usepackage[dvipsnames]{xcolor}

\definecolor{cvprblue}{rgb}{0.21,0.49,0.74}
\usepackage[pagebackref,breaklinks,colorlinks,citecolor=cvprblue]{hyperref}

\title{Multi-Human Mesh Recovery with Transformers}

\author{Zeyu Wang, Zhenzhen Weng, Serena Yeung-Levy\\
Stanford University\\
{\tt\small \{wangzeyu,zzweng,syyeung\}@stanford.edu}
}

\begin{document}
\maketitle
\begin{abstract}

    Conventional approaches to human mesh recovery predominantly employ a region-based strategy. This involves initially cropping out a human-centered region as a preprocessing step, with subsequent modeling focused on this zoomed-in image. While effective for single figures, this pipeline poses challenges when dealing with images featuring multiple individuals, as different people are processed separately, often leading to inaccuracies in relative positioning.  Despite the advantages of adopting a whole-image-based approach to address this limitation, early efforts in this direction have fallen short in performance compared to recent region-based methods. In this work, we advocate for this under-explored area of modeling all people at once, emphasizing its potential for improved accuracy in multi-person scenarios through considering all individuals simultaneously and leveraging the overall context and interactions. We introduce a new model with a streamlined transformer-based design, featuring three critical design choices: multi-scale feature incorporation, focused attention mechanisms, and relative joint supervision. Our proposed model demonstrates a significant performance improvement, surpassing state-of-the-art region-based and whole-image-based methods on various benchmarks involving multiple individuals.
    
\end{abstract}    
\section{Introduction}
\label{sec:intro}

    Monocular Human Mesh Recovery (HMR) has drawn considerable recent attention owing to its wide array of applications, spanning from augmented reality to computer assisted coaching~\cite{tian2023recovering}. Various methods for human mesh recovery have been introduced over the years~\cite{kanazawa2018end,li2022cliff,ROMP,BEV,kocabas2020vibe,kocabas2021pare}. Models that employ a straightforward end-to-end training mechanism to directly estimate SMPL \cite{loper2023smpl} parameters are known as regression-based models~\cite{kanazawa2018end}. They stand in contrast to optimization-based methods \cite{bogo2016keep} that fine-tune the SMPL parameter estimates in an iterative optimization cycle. Notably, regression-based approaches sidestep certain challenges associated with optimization-based methods, such as getting trapped in local minima. As a result, most leading-edge human mesh recovery models adopt a regression-based approach~\cite{li2022cliff,goel2023humans,wang2023refit}.

\begin{figure}[t]
  \centering
   \includegraphics[width=1.0\linewidth]{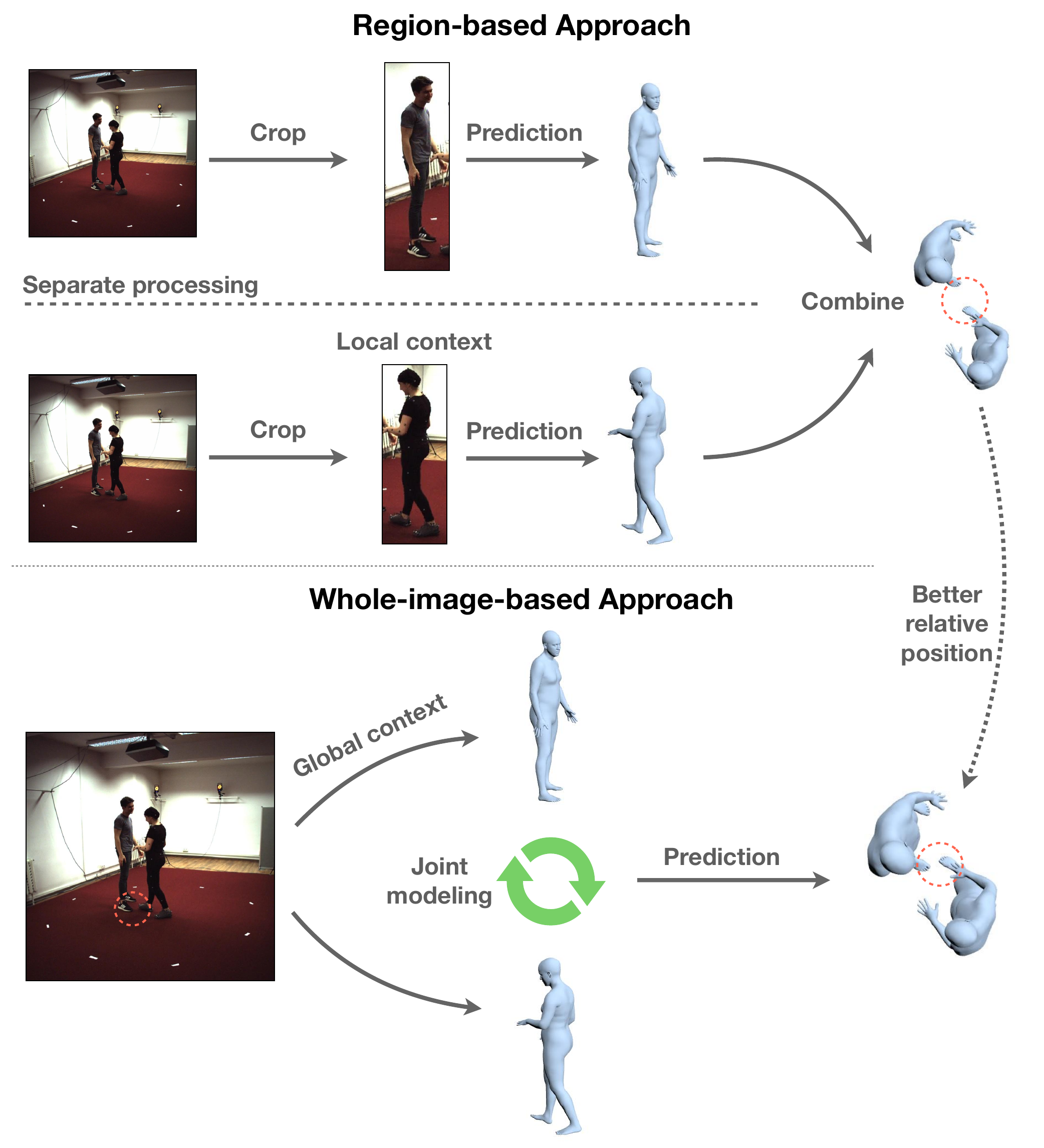}

   \caption{(Top) The majority of current human mesh recovery techniques adopt a region-based approach. They begin by isolating human-centric regions and processing them individually. This often results in inaccuracies in the relative positioning of multiple individuals. (Bottom) We advocate for the adoption of whole-image-based approach, where all people are processed simultaneously. This direction is less explored and early works are outperformed by recent region-based methods by a large margin. In this work, we introduce a new model that incorporates several crucial design choices, showcasing a substantial improvement in accurately modeling all individuals and surpassing the performance of state-of-the-art region-based models.}
   
   \label{fig:pull}
\end{figure}

    Despite their seemingly simplistic architectural design, most regression-based methods rely on local input information to predict SMPL parameters and necessitate a pre-processing step to isolate the region of interest~\cite{kanazawa2018end}. To be specific, from a given input image, the individual of interest is initially cropped based on a bounding box around them. This cropped local region, with the person centered, is then input into the network to deduce the SMPL parameters for that individual. The logic behind this approach is to discard any extraneous information unrelated to the person, allowing the model to concentrate solely on the subject of interest. While this strategy is logical for images with just one individual, it presents complications when multiple people are depicted in the same frame~\cite{yin2023hi4d,patel2021agora}, which is more common for many practical use cases.

    Firstly, when predicting for an image containing multiple people, each region has to be processed by the network individually, leading to a computational scaling that is directly proportional to the number of people present in the image. Even more importantly, by isolating each region and modeling individuals independently, region-based methods overlook the global context of multiple people in a scene, thereby missing the chance to process everyone collectively and utilize the relative information among them. This fundamental limitation of region-based approaches hinders their capability to precisely represent the relative positioning of multiple individuals in a scene (as illustrated in top of Figure~\ref{fig:pull}). Furthermore, this problem is not readily apparent, as most standard HMR benchmarks primarily concentrate on single-person scenarios~\cite{von2018recovering}, and the commonly-used error metrics are computed individually for each person. 

    Given this inherent limitation of region-based methods, we advocate for adopting a whole-image-based approach to address scenes with multiple people. This approach processes the entire input image in one go, generating meshes for all individuals within the image simultaneously. Although this approach holds promise, it currently enjoys less popularity and remains relatively under-explored, as recent advancements in region-based approaches have outperformed early attempts~\cite{ROMP,BEV,fieraru2021remips} by a significant margin. The reason behind this lies in the unique challenge faced by whole-image-based methods: maintaining detailed subject information while filtering out irrelevant background information, a problem circumvented effectively by region-based methods through the cropping preprocessing step.

    To overcome this challenge, we propose a novel whole-image-based model characterized by a streamlined transformer architecture design. This model leverages multi-scale features to retain both high-level context and crucial low-level details necessary for precise localization. Furthermore, it employs a reduced-sized attention module to discard irrelevant background information and focus on modeling the human subjects. Additionally, to fully harness the advantages of processing multiple people simultaneously, we introduce a novel relative joint loss function, specifically designed to supervise relative joint locations. By combining these techniques, our proposed whole-image-based model significantly outperforms state-of-the-art region-based methods on various benchmarks containing scenes with multiple people. We hope our work will draw interest in whole-image-based methods and encourage further exploration in this area.
    
    In summary, we make the following contributions:

    \begin{itemize}

        \item We advocate for the adoption of whole-image-based HMR methods, an area that has received less exploration compared to the prevalent region-based methods. This approach proves effective in leveraging the global context, addressing the inherent limitation of the latter.

        \item We introduce a new whole-image-based method featuring a streamlined transformer-based design. It addresses challenges encountered by previous methods by efficiently utilizing multi-scale features and a focused attention mechanism.

        \item We introduce a novel relative joint loss, designed specifically to supervise relative joint locations, maximizing the benefits of processing multiple individuals simultaneously.

        \item Experiments on various multi-human benchmarks reveal that our method surpasses both existing whole-image-based and region-based methods significantly. This is particularly evident in the joint evaluation metric, which assesses the relative positioning of all subjects in a scene.
    
    \end{itemize}

\section{Related Work}
\label{sec:related-work}

\paragraph{Single-view Human Mesh Recovery.} Human mesh recovery (HMR) involves the task of estimating human pose and shape with a target output in the mesh format~\cite{pavlakos2022human,tian2023recovering,Pavlakos2018LearningTE,Baradel2021LeveragingMD,Lin2020EndtoEndHP,Kolotouros2021ProbabilisticMF,Xu2019DenseRaCJ3,Kolotouros2019ConvolutionalMR,Kolotouros2019LearningTR,Omran2018NeuralBF}. Specifically, single-view HMR focuses on generating meshes from a single input image. This capability is highly valuable in various practical applications and offers a promising alternative to the cumbersome motion capture systems~\cite{Rajasegaran2022TrackingPB,Ye2023DecouplingHA,yuan2022glamr,Lin2021MeshG,Mller2023GenerativePA}. The introduction of SMPL~\cite{loper2023smpl}, a parametric representation of the human body, has led to a shift in the task's approach. Many studies now aim to estimate SMPL parameters from input images. These approaches can be broadly classified into two groups. Optimization-based methods refine the SMPL parameter estimates through iterative processes~\cite{bogo2016keep}, while regression-based models directly predict SMPL parameters in an end-to-end manner~\cite{kanazawa2018end}. Notably, regression-based techniques circumvent certain challenges associated with optimization-based methods, such as the risk of getting stuck in local minima. Consequently, most recent research adopts a regression-based approach~\cite{BEV,ROMP,li2022cliff,goel2023humans,wang2023refit}, and our work falls into this category as well.

\paragraph{Region-Based vs. Whole-Image-Based HMR.} Since Kanazawa et al.'s groundbreaking work~\cite{kanazawa2018end}, which introduced an end-to-end model for directly regressing shape and pose parameters from image pixels without intermediate 2D keypoint detection, most regression-based Human Mesh Recovery (HMR) methods have adopted a region-based approach~\cite{li2022cliff,wang2023refit}. This approach involves using cropping as a preprocessing step to isolate human regions, with the model focusing solely on these human-centered areas for predictions. Most recent advancements in this area include CLIFF~\cite{li2022cliff}, HMR2.0~\cite{goel2023humans}, ReFIT~\cite{wang2023refit}, among others~\cite{zhang2023pymaf,zhang2021pymaf,kocabas2021pare,kocabas2020vibe}. However, a major drawback of region-based methods is their inefficiency and potential inaccuracy in positioning when dealing with images containing multiple people, as each individual must be processed separately. In contrast, whole-image-based methods process the entire image at once, generating meshes for all individuals simultaneously, but this approach remains underexplored as early attempts~\cite{ROMP,BEV,fieraru2021remips} have been significantly outperformed by recent region-based methods~\cite{li2022cliff,goel2023humans}. In our work, we introduce a new model with a streamlined transformer-based design. Through several critical design choices, our model shows improved performance over both previous whole-image-based and region-based methods. 

\paragraph{Transformer Models.} Ever since the seminal work of vision transformer~\cite{dosovitskiy2020image}, which successfully introduced the transformer model~\cite{vaswani2017attention} widely used in natural language processing (NLP) to the computer vision domain, researchers have consistently invested efforts in exploring various adaptations of the standard attention mechanism tailored to diverse vision tasks, each with its unique inductive bias~\cite{liu2021swin,liu2022video,dong2022cswin,yang2021focal,Rao2021DynamicViTEV,Tu2022MaxViTMV,xia2022vision}. In this work, we leverage the deformable attention module initially proposed by Zhu et al.~\cite{zhu2020deformable}, designed originally to enhance performance and convergence speed in transformer-based object-detection models~\cite{carion2020end,zhang2022dino}. We demonstrate the critical role of the focused attention mechanism provided by the deformable attention module in our whole-image-based approach. This mechanism allows our method to concentrate on pertinent information while disregarding irrelevant background content devoid of people. As far as the authors are aware, this specific application of deformable attention has not been employed in the context of HMR tasks before.

\section{Method}
\label{sec:method}

In this section, we first introduce the preliminaries of statistical human body model, which is the cornerstone for regression-based HMR methods, including our work. Following this, we will delve into the detailed architecture of our proposed method for multi-person mesh recovery from whole images. Lastly, we will outline the loss function employed in training our model.

\begin{figure*}[t]
  \centering
  \includegraphics[width=0.95\linewidth]{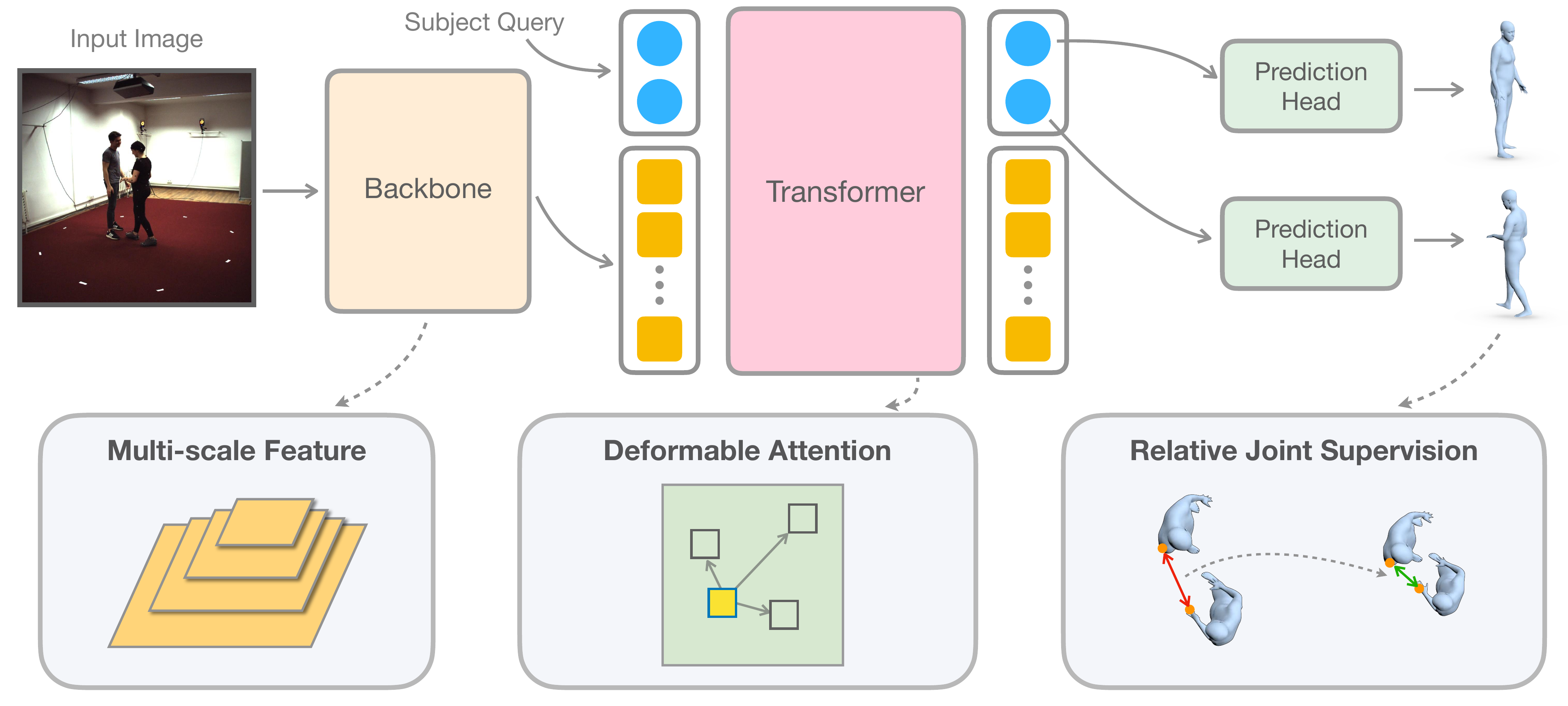}

   \caption{Our proposed approach adopts a streamlined transformation-based design, processing the entire image and generating meshes for all individuals simultaneously. It incorporates three crucial design elements that effectively prioritize essential regions and model the relative positions of humans, ultimately leading to improved performance compared to existing methods in both region-based and whole-image-based approaches.}
   \label{fig:model}
\end{figure*}

\subsection{Preliminaries}

    \paragraph{Body Model.} Introduced by \citet{loper2023smpl} SMPL (Skinned Multi-Person Linear) model offers a parametric representation of the human body, characterized by its capacity to capture the intricate details of human anatomy, skeletal structure, and skin deformation. It represents the human body as a deformable mesh through a linear combination of shape and pose parameters. Mathematically, SMPL is parameterized with pose parameters $\theta \in \mathbb{R}^{24 \times 3 \times 3}$ and shape parameters $\beta \in \mathbb{R}^{10}$. It outputs a human mesh $\mathcal{M} \in \mathbb{R}^{3 \times N}$, with $N=6890$ vertices. And body joints $\mathbf{X} \in \mathbb{R}^{3 \times K}$ can be regressed with a linear combination of mesh vertices, \ie $\mathbf{X} = \mathcal{M}\mathcal{J}$, where $\mathcal{J} \in \mathbb{R}^{N \times K}$ is the regressing weights. Following the same design methodology, the follow-up work SMPL-X \cite{pavlakos2019expressive} extends SMPL with a higher-dimensional pose space, allowing for greater flexibility in modeling complex and expressive human body configurations, encompassing aspects such as hand gestures and facial expressions.

\subsection{Architecture}

    Figure~\ref{fig:model} provides an overview of our proposed model's architecture, which takes a streamlined transformer-based design. It begins by taking an input image, which undergoes initial processing through a backbone network to extract feature maps. These feature maps are then divided and converted into a series of tokens, supplemented with positional embeddings to encode their spatial information. The transformer encoder subsequently processes these tokens, facilitating information exchange among them. Then on the decoder side, a collection of queries is introduced, each representing a candidate individual. In the decoding layers, these query tokens engage with both themselves and the output tokens from the transformer encoder. After several decoding layers, each query is processed by a shared prediction head that produces 
    predictions for a range of parameters, including pose parameters, shape parameters, and translation values. This translation information indicates the relative positioning of the individual in relation to the camera.

    One significant challenge in whole-image-based human reconstruction methods is the presence of areas in the full image without any human elements, offering no relevant information for the task. In contrast, region-based models circumvent this issue by isolating individual subjects with a separate cropping step. The effectiveness of whole-image-based methods, therefore, hinges on the ability to discern and utilize pertinent information while disregarding irrelevant data. To overcome this challenge, we employ two specific techniques, which are fundamental to the effectiveness of the proposed model. 
    
    \paragraph{Multi-scale Features.} In scenarios where the main subject within the input image occupies a limited portion, it becomes essential to preserve crucial image details in the early stages. This ensures that later stages have access to the necessary information for accurately locating joints. Our proposal involves the utilization of multi-scale features~\cite{lin2017feature}, where not only the final output but also the outputs of early layers of the backbone are forwarded to subsequent stages for further processing. This hierarchical grouping of features preserves the overall contextual information about the person's location within the scene, while also capturing detailed information about local joints. This dual focus is essential for accurately estimating pose parameters. 
    
    \paragraph{Deformable Attention Module.} For transformer encoder and decoder, we adopt the deformable attention technique as introduced by \citet{zhu2020deformable}, which differs from the standard transformer model's full self-attention mechanism~\cite{vaswani2017attention,dosovitskiy2020image}. The key aspect of this method is that attention is focused only on a limited number of sampling points, each with an offset relative to a reference location, thereby optimizing the attention process. In the subsequent experiment, we demonstrate that the constrained size of the attention mechanism plays a crucial role in enabling the network to disregard irrelevant information across the entire image where no person is present. This allows the model to concentrate its focus on the specific task of modeling humans. 
    
    Specifically, let $x \in \mathbb{R}^{C\times H \times W}$ represents the input feature map, $z_q$ represents the query feature, then the deformable attention is calculated as (for simplicity, we omit the multi-head attention):

    \begin{equation}
        \text{DeformAttn}(z_q, p_q, x) = \sum_{k=1}^K A_{k} \cdot W x(p_q + \Delta p_{k})
    \end{equation}

    Here, k denotes the number of sampling points, which is far less than the number of total feature tokens ($K \ll H \times W$). $p_q$ is the reference point, and $\Delta p_k$ is the sampling offset. Both the attention weight $A_k$ and the sampling offset $\Delta p_k$ are calculated with a linear projection of the input query feature $z_q$. 

\subsection{Loss Functions}

    Following previous works on human mesh recovery~\cite{li2022cliff,goel2023humans}, we train the proposed model with a combination of joint losses in both 2D and 3D, and SMPL parameter losses. Specifically, the 3D keypoint loss is calculated as following:

        \begin{equation}
            \mathcal{L}_{3D} = \| \mathbf{X} - \mathbf{X}^* \|_1
        \end{equation} 
        
    where $\mathbf{X}$ is the predicted keypoints in 3D and $\mathbf{X}^*$ is the corresponding grountruth 3D keypoints. Additional 2D supervision is added by projecting the predicted 3D joints onto the camera frame, and comparing it with groundtruth with an L1 loss:

        \begin{equation}
            \mathcal{L}_{2D} = \| \pi(\mathbf{X}) - \mathbf{x}^* \|_1
        \end{equation}
        
    where $\pi$ is the prospective projection function and $\mathbf{x}^*$ is the groundtruth 2D keypoints.
    
    In addition, when the groundtruth SMPL parameters are available, we add direct supervision on these parameters:

        \begin{equation}
            \mathcal{L}_{SMPL} = \| \theta - \theta^* \|_1 + \| \beta - \beta^* \|_1
        \end{equation}

    where, $\theta$ and $\beta$ are the pose and shape parameters of SMPL, and $\theta^*$, $\beta^*$ are the corresponding groundtruth. 

    \paragraph{Relative Joint Loss.} A significant advantage of whole-image-based methods over region-based approaches is their ability to model multiple individuals simultaneously. This process begins at the architectural level, where queries from different candidates interact and share information during the attention process in the decoder layers. Notably, the concept of joint modeling can also be leveraged in the loss function. In addition to the joint loss that supervises the individual positions of each person, we can introduce extra supervision focusing on the relative positions of joints among multiple humans.
    
    Building on this insight, we propose a novel relative joint loss to effectively utilize this information. Specifically, let $\mathbf{\hat{X}} \in \mathbb{R}^{3 \times nK}$ denote the 3D location of the joints for all individuals, where $n$ is the number of individuals, and each individual has $K$ joints. Then the relative displacement among all the joints of all people is represented as a matrix $\overrightarrow{\mathbf{D}} \in \mathbb{R}^{3 \times nK \times nK}$, where each element of the last two dimension is the relative displacement of the two corresponding joints. Denoting the groundtruth displacement matrix as $\overrightarrow{\mathbf{D}}^*$, then the proposed relative joint loss contains two components that align the relative distance and direction between the predicted and the target relative displacement. Specifically, the relative distance loss is calculated as,

        \begin{equation}
            \mathcal{L}_{rt-distance} = \| |\overrightarrow{\mathbf{D}}| - |\overrightarrow{\mathbf{D}}^*|  \|_1
        \end{equation}
        
    where $|\overrightarrow{\mathbf{D}}| \in \mathbb{R}^{nK \times nK}$ is the the norm of the 3D displacement. And the relative directional loss is,

        \begin{equation}
            \mathcal{L}_{rt-directional} = 1 - \frac{1}{(nK)^2} \sum_{i,j} \frac{\overrightarrow{\mathbf{D}}_{\cdot ij}}{\| \overrightarrow{\mathbf{D}}_{\cdot ij} \|} \cdot 
             \frac{\overrightarrow{\mathbf{D}}^*_{\cdot ij}}{\| \overrightarrow{\mathbf{D}}^*_{\cdot ij} \|}
        \end{equation}
        
    which maximizes the cosine similarity between the predicted and target displacement matrix. 

    The overall combined loss is:

    \begin{equation}
        \begin{split}
            \mathcal{L} &= \lambda_{3D}\mathcal{L}_{3D} + \lambda_{2D}\mathcal{L}_{2D}  \\
            &\quad + \lambda_{SMPL}\mathcal{L}_{SMPL} + \lambda_{rt}\mathcal{L}_{rt} 
        \end{split}
    \end{equation}

    where the relative joint loss is a combination of two elements $\mathcal{L}_{rt} = \mathcal{L}_{rt-directional} + \lambda \mathcal{L}_{rt-directional}$ and $\lambda$'s are hyper-parameters controlling relative importance of each loss term.

\section{Experiments}
\label{sec:experiments}

In this section, we begin by presenting the experimental setup (Section \ref{subsec:experiment_setup}), which encompasses details regarding the datasets used, the evaluation metrics employed, and specific implementation details. Subsequently, we provide the quantitative results (Section \ref{subsec:quant_results}), followed by the ablation studies (Section \ref{subsec:ablation_study}) and qualitative results.

\subsection{Experiment Setup}
\label{subsec:experiment_setup}
\paragraph{Datasets.} We experiment with three datasets with presence of multiple people in each image:

    \begin{itemize}
        \item \textbf{CHI3D}. Close Interactions 3D (CHI3D) \cite{fieraru2020three} is an indoor Motion Capture dataset featuring closely interacting dyads. The dataset contains multiple pairs of subjects performing different actions (e.g. hug, posing). We use two subject pairs for training and one subject paris for test.

        \item \textbf{Hi4D}. 4D Instance Segmentation of Close Human Interaction (Hi4D) \cite{yin2023hi4d} by Yin \etal proposes a method to utilize individually fitted avatars and an alternating optimization scheme for segmenting fused raw scans into individual instances. A high-quality dataset of close-interacting humans is  then compiled featuring 20 subject pairs, and more than 11K frames. We utilize ten pairs for training and five pairs for validation and test respectively.

        \item \textbf{BEDLAM}. Bodies Exhibiting Detailed Lifelike Animated Motion (BEDLAM) \cite{black2023bedlam} is a recent large-scale multi-person synthetic dataest featuring diverse body shapes, motions, and realistic clothing. We select one of the indoor scenes for training and test.
        
    \end{itemize}

\paragraph{Evaluation.}

    For evaluation, we employ standard MPJPE, PA-MPJPE, and PVE metrics. MPJPE, or Mean Per Joint Position Error, calculates the average L2 error across all joints after alignment to the root node. PA-MPJPE offers a similar calculation but is determined after aligning the predicted pose to the ground-truth pose using Procrustes Alignment. Meanwhile, PVE, which represents the per-vertex error, provides a more detailed assessment. In addition to the three standard metrics, we incorporate joint PA-MPJPE (J-PA-MPJPE). It is analogous to PA-MPJPE but are computed collectively for all individuals present in the image, which captures the relative positions of people within the scene and is essential for assessing whether the reconstructed human meshes maintain their respective global positions.
    
\paragraph{Implementation details.}

    We employ ResNet-50 \cite{he2016identity} as the backbone and extract features from the outputs after the second, third, and fourth blocks to construct multi-scale feature maps. The parameters of the deformable transformer encoder are shared across various feature levels. Our model consists of six attention layers for both the encoder and decoder components, with eight attention heads and four sampling points for the deformable attention mechanism. During the encoding stage, the positional information of each token within the 2D feature map is utilized as the reference position for attention sampling. At the decoding stage, the query is generated with a linear projection of the positional encoding of the corresponding individual and the 2D position of each individual serves as the reference point for attention. We use AdamW \cite{loshchilov2017decoupled} as the optimizer with a learning rate of 5e-5.

\paragraph{Baselines.} 

    We compare the proposed model with both the region-based and whole-image-based approaches.  Among the region-based methods, we consider CLIFF~\cite{li2022cliff}, which extends the original HMR~\cite{kanazawa2018end} method by incorporating bounding box information alongside cropped image features, enhancing its ability to predict global rotation in the original camera frame. We still categorize it as region-based method as it does not rely on whole-image feature maps and makes predictions for each individual independently. Additionally, we include HMR2.0~\cite{goel2023humans}, a recent approach that employs a simplified transformer architecture for mesh recovery. Both CLIFF and HMR2.0 represent state-of-the-art methods for this task. For the whole-image-based approaches, we compare with ROMP~\cite{ROMP}, which directly generate multi-person mesh in a single-stage fashion, utilizing a body center heatmap and associated parameter maps. And a follow-up work BEV~\cite{BEV} that extends ROMP with an additional imaginary bird's-eye-view representations to explicitly reason about relative depth of multiple individuals. We use the public available implementation of CLIFF from \cite{black2023bedlam} and official implementations provided by the original authors for the other three baselines.

\subsection{Experiment Results}
\label{subsec:quant_results}
    \subsubsection{Results on CHI3D}

\begin{table}[t]
  \centering
  \resizebox{\columnwidth}{!}{%
  \begin{tabular}{lcccc}
    \toprule
    Method & MPJPE & PA-MPJPE & PVE &  J-PA-MPJPE \\
    \midrule
    ROMP & 113.5 & 87.0 & 143.2 & 222.5 \\
    BEV & 69.8 & 56.0 & 86.8 & 92.9  \\
    \midrule
    HMR2.0 & 64.8 & 37.7 & 83.8 & 87.3 \\
    CLIFF & 55.8 & 38.7 & 77.2 & 81.6  \\
    \midrule
    Ours & \textbf{50.4} & \textbf{35.5} & \textbf{71.1} & \textbf{52.8} \\
    \bottomrule
  \end{tabular}}
  \caption{Comparison of the proposed method with other region-based and whole-image based methods on CHI3D dataset. J-PA-MPJPE stands for joint PA-MPJPE. HMR2.0 and CLIFF are region-based methods. ROMP, BEV and ours are whole-image-based methods. Our model surpasses others in performance, notably in the joint evaluation metric. This enhanced performance is attributed to its superior ability to model the relative positions of humans, effectively leveraging full-image context through advanced attention mechanisms.
  }
  \label{tab:chi3d}
\end{table}

        Table~\ref{tab:chi3d} shows the quantitative comparison of all methods on the CHI3D dataset, and our proposed model outperforms all other baselines with a large margin. To begin with, notice that both region-based methods, HMR2.0 and CLIFF, achieve fairly good performance in modeling human pose, with PA-MPJPE under 40. Comparing the two, while HMR2.0 achieves a slight higher PA-MPJPE of 37.7 over CLIFF's 38.7, it lags behind to CLIFF on other metrics, scoring 64.8 on MPJPE and 83.8 on PVE against CLIFF's 55.8 and 77.2, respectively. This might be due to HMR2.0's use of a large backbone (ViT Huge)~\cite{dosovitskiy2020image}, which could be less sample efficient with smaller datasets compared to CLIFF's convolutional backbone. It is important to notice the disparity in scores between the joint evaluation metric J-PA-MPJPE and its individual counterpart, PA-MPJPE, for both methods is significant (87.3 vs. 37.7 for HMR2.0 and 81.6 vs. 38.7 for CLIFF). This discrepancy underscores the inherient limitation of region-based methods for accurately modeling relative positions of multiple people.

        Regarding the whole-image-based baselines, ROMP and BEV,  despite their theoretical advantage in using contextual information to model the relative positions of interacting humans, they don't outperform the more recent region-based methods. BEV, for instance, records an MPJPE of 69.8, PA-MPJPE of 56.0, and PVE of 86.8 (compared to CLIFF's 55.8, 38.7 and 77.2). This illustrates that it is non-trivial for whole-image-based methods to perform well and surpass the advantages offered by the strong prior introduced by cropping out the individual of interest in region-based methods.

        Finally, with a careful design, our proposed model achieves the best performance, surpassing other whole-image-based methods as well as the strong region-based models across various metrics, with MPJPE of 50,4 PA-MPJPE of 35.5, PVE of 71.1 and joint PA-MPJPE of 52.8. Notably, in comparison to other baselines, our model not only shows significant improvement in standard metrics (50.4 vs. 55.8, 35.5 vs. 37.7, 71.1 vs. 83.8 for MPJPE, PA-MPJPE, and PVE, compared to the top-performing baseline) but also demonstrates an even more substantial enhancement in the joint PA-MPJPE (52.8 vs. 81.6). This demonstrates that our approach effectively harnesses the advantages of the full-image context. By modeling multiple individuals collectively, it captures the relative positions of multiple people in the scene more accurately, addressing the inherent limitations of region-based methods.
        
    \subsubsection{Results on Hi4D}

\begin{table}[t]
  \centering
  \resizebox{\columnwidth}{!}{%
  \begin{tabular}{lcccc}
    \toprule
    Method & MPJPE & PA-MPJPE & PVE &  J-PA-MPJPE \\
    \midrule

    ROMP & 167.2 & 103.5 & 215.3 & 221.1 \\
    BEV & 120.9 & 96.8 & 153.9 & 154.9 \\

    \midrule

    HMR2.0 &  108.6 & 80.3 & 141.2 & 144.1\\
    CLIFF & 106.4 & 81.3 & 138.8 & 132.5 \\
    
    \midrule
    Ours & \textbf{102.1} & \textbf{75.2} & \textbf{132.7} & \textbf{105.6}\\
    \bottomrule
  \end{tabular}
  }
  \caption{Comparison of the proposed method with other region-based and whole-image based methods on Hi4D dataset.}
  \label{tab:hi4d}
\end{table}

        Table~\ref{tab:hi4d} presents a comparison of our proposed model with other region-based and whole-image-based methods on the Hi4D dataset. First observe that the performance of all methods on the Hi4D dataset is lower compared to the CHI3D dataset. This decrease in performance can be attributed to the Hi4D dataset containing more subject pairs and a wider variety of close human interactions than the CHI3D dataset. On this more challenging dataset, our proposed method still manages to outperform other baselines by a large margin. It achieves MPJPE of 102.1, PA-MPJPE of 75.2, PVE of 132.7 and joint PA-MPJPE of 105.6. This represents an enhancement of 4.3 points in MPJPE, 5.1 points in PA-MPJPE, 6.1 points in PVE, and 26.9 points in joint PA-MPJPE over the top-performing baseline. This outcome further confirms the superior capability of our proposed method in accurately modeling multiple humans.

    \subsubsection{Results on BEDLAM}

        Performance of different models on the BEDLAM dataset is summarized in Table~\ref{tab:bedlam}. Unlike CHI3D and Hi4D, BEDLAM features scenes with more than two individuals. However, each person's movements are independently sampled from the AMASS \cite{mahmood2019amass} dataset and do not interact, resulting in a greater spatial separation between them. This gives benefits to the region-based models, and narrows the performance gap between our model and CLIFF on the PA-MPJPE metric (50.8 vs. 51.0), which specifically assesses pose accuracy. However, when evaluating on the other metrics that take into account global orientation and translation, our method still outperforms by a large margin. It surpasses the best-performing baseline by 6.2 points in MPJPE, by 6.5 points in PVE, and most notably, by 108.5 points in joint PA-MPJPE. The significant gain in J-PA-MPJPE, compared to approximately 30-point improvement seen in CHI3D and Hi4D datasets, can also be ascribed to the presence of more individuals and their increased spatial separation.

\begin{table}[t]
  \centering
  \resizebox{\columnwidth}{!}{%
  \begin{tabular}{lcccc}
    \toprule
    Method & MPJPE & PA-MPJPE & PVE &  J-PA-MPJPE \\
    \midrule

    ROMP & 169.3 & 122.2 & 218.7 & 431.2 \\
    BEV &  97.9 & 75.3 & 124.3 & 216.3 \\
    
    \midrule
    
    HMR2.0 & 90.9 & 69.3 & 116.2 & 274.0 \\
    CLIFF & 73.8 & 51.0 & 97.1 & 206.7  \\
    
    \midrule
    Ours & \textbf{67.6} & \textbf{50.8} & \textbf{90.6} & \textbf{98.2}\\
    \bottomrule
  \end{tabular}
  }
  \caption{Comparison of the proposed method with other region-based and whole-image based methods on BEDLAM dataset.}
  \label{tab:bedlam}
\end{table}

\subsection{Ablation Study}
\label{subsec:ablation_study}

\begin{table}[t]
  \centering
  \small
  \resizebox{\columnwidth}{!}{%
  \begin{tabular}{lcccc}
    \toprule
    Method & MPJPE & PA-MPJPE & PVE &  J-PA-MPJPE \\
    \midrule
    Ours & \textbf{67.6} & \textbf{50.8} & \textbf{90.6} & \textbf{98.2} \\
    \midrule
    w/o Relative-loss & 69.1 & 52.0 & 92.6 & 102.1\\
    w/o Multi-scale & 70.9 & 52.7 & 95.0 & 102.4\\
    w/o DeformAtt & 74.5 & 54.8 & 98.2 & 109.4\\
    \bottomrule
  \end{tabular}}
  \caption{Ablation results on BEDLAM dataset. This highlights the significance of three essential design elements: relative joints loss, multi-scale features and deformable attention. These components are crucial for the proposed model's ability to effectively process whole-image input and exceed the performance of strong region-based as well as whole-image-based baselines.}
  \label{tab:ablation}
\end{table}

    We run ablation study to show the importance of the key design choices of our proposed model and the results are summarized in table~\ref{tab:ablation}. The absence of relative joint loss results in a noticeable decrease in performance: a 1.5-point increase in MPJPE, a 1.2-point increase in PA-MPJPE, a 2.0-point increase in PVE, and a 3.9-point increase in joint PA-MPJPE. Observe that the joint PA-MPJPE metric experiences a larger increase compared to other measures, indicating that the loss more effectively improves the relative positioning of multiple individuals. In the absence of multi-scale features, there is an additional increase of 1.8, 0.7, 2.4, and 0.3 points in MPJPE, PA-MPJPE, PVE, and joint PA-MPJPE metrics, respectively. The more significant performance deterioration in MPJPE and PVE compared to PA-MPJPE and joint PA-MPJPE indicates that multi-scale features are particularly crucial for accurately capturing precise joint locations, which depend heavily on detailed low-level features. The performance suffers more dramatically when deformable attention is substituted with standard attention. This change leads to further increase of 3.6 points in MPJPE, 2.1 points in PA-MPJPE, 3.2 points in PVE, and 7.0 points in joint PA-MPJPE, rendering its performance inferior to that of the best-performing baselines. This significant drop underscores the vital role of deformable attention in enabling the model to disregard irrelevant image regions and concentrate more effectively on important information.

\begin{figure*}[t]
  \centering
 \includegraphics[width=1.0\linewidth]{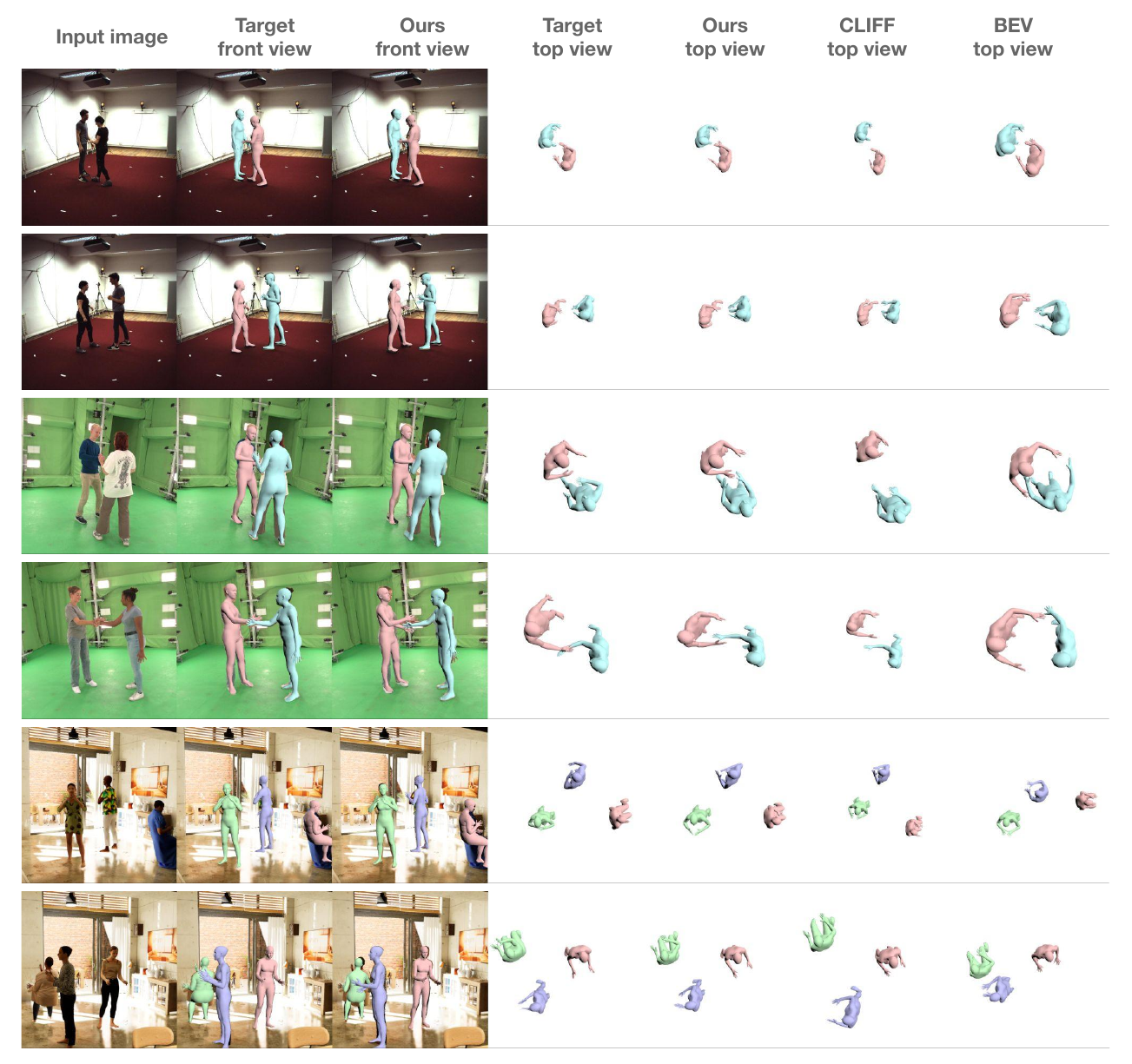}

   \caption{Qualitative comparisons on CHI3D~\cite{fieraru2020three} (top 2 rows), Hi4D~\cite{yin2023hi4d} (middle 2 rows), and BEDLAM~\cite{black2023bedlam} (bottom 2 rows). Meshes predicted by our method are overall more accurate in terms of the relative locations and orientations compared to top-performing region-based and whole-image-based baselines.}

   \label{fig:qualitative}
\end{figure*}

\subsection{Limitations and Future Work}
\label{subsec:limitations}

While our approach demonstrates improved accuracy in determining the global positioning and relative orientation of predicted human meshes, it still faces a common limitation of regression-based methods: the possibility of mesh penetration when people are in close interaction. We believe whole-image-based approaches like ours hold more chance to address the problem compared to region-based methods. Future work could consider augmenting our method with contact optimization strategies to enhance the overall fidelity of the reconstructions in such scenarios. 

\section{Conclusion}
\label{sec:conclusion}

In this work, we delve into the relatively unexplored domain of simultaneous multi-person modeling, underscoring its potential to enhance accuracy in multi-individual scenarios. We introduce a novel model with a streamlined transformer-based architecture, which incorporates three pivotal design choices: the integration of multi-scale features, focused attention mechanisms, and relative joint supervision. The results of our proposed model show a notable performance boost, surpassing the capabilities of both state-of-the-art region-based and whole-image-based methods across diverse benchmarks that involve multiple individuals.

\clearpage

{
    \small
    \bibliographystyle{ieeenat_fullname}
    \bibliography{main}
}


\end{document}